\documentclass[12pt]{article}
\usepackage{graphicx}
\usepackage{chicago}
\usepackage{geometry}
\usepackage{siunitx} 
\title{Optimizing the optimizer for data driven deep neural networks and physics informed neural networks} 
\author{John Taylor\,$^{1,2,3,*}$, Wenyi Wang\,$^{3}$,  Biswajit Bala\,$^{3}$, Tomasz Bednarz\,$^{1}$}

\begin{document}

\maketitle

\noindent
\textsuperscript{1}CSIRO Data61, Canberra, Australian Capital Territory, Australia\\
\textsuperscript{2}College of Engineering and Computer Science, The Australian National University, Canberra, Australia\\
\textsuperscript{3}Defence Science and Technology Group, Department of Defence, Canberra Australia\\

\textsuperscript{*}Corresponding Author Address: John Ashley Taylor, CSIRO Data61, CSIRO Synergy Building, Clunies Ross Street, Black Mountain, Canberra ACT 2601.\\
E-mail: John.Taylor@data61.csiro.au

\pagebreak

\section*{Abstract}

We investigate the role of the optimizer in determining the quality of the model fit for neural networks with a small to medium number of parameters. We study the performance of Adam, an algorithm for first-order gradient-based optimization that uses adaptive momentum, the Levenberg-Marquardt (LM) algorithm a second order method, Broyden–Fletcher–Goldfarb–Shanno algorithm (BFGS) a second order method and L-BFGS, a low memory version of BFGS. Using these optimizers we fit the function $y = sinc(10x)$ using a neural network with a few parameters. This function has a variable amplitude and a constant frequency. We observe that the higher amplitude components of the function are fitted first and the Adam, BFGS and L-BFGS struggle to fit the lower amplitude components of the function.  We also solve the Burgers equation using a physics informed neural network(PINN) with the BFGS and LM optimizers. For our example problems with a small to medium number of weights, we find that the LM algorithm is able to rapidly converge to machine precision offering significant benefits over other optimizers. We further investigated the Adam optimizer with a range of models and found that Adam optimiser requires much deeper models with large numbers of hidden units containing up to $\approx$26x more parameters, in order to achieve a model fit close that achieved by the LM optimizer. The LM optimizer results illustrate that it may be possible build models with far fewer parameters. We have implemented all our methods in Keras and TensorFlow 2.


\section{Introduction}

Neural networks have been successfully applied to a broad range of applications and are now increasingly employed to gain new insights into complex and demanding science problems~\cite{Lecun2015}. Many science problems, in the context of deep learning, are considered as regression problems where obtaining the best model fit is a key measure of success.  The best model fit is typically the result of applying an optimisation method that minimises a loss function.  Improvements in the model fit are often a key criteria used to establish that a new model delivers improved performance over prior modelling approaches. 

Training neural networks usually continues until the optimizer achieves a minimum value where the loss function no longer decreases or where a comparison of the loss achieved on the training and test data indicates that over-fitting is taking place. The optimizer may not succeed and cannot guarantee that the global minimum has been achieved. The selection and application of the optimizer therefore warrants careful consideration. 

Neural networks are considered to be universal approximators~\cite{Hornik1989}. While this property gives neural networks incredible flexibility, as evidenced by their application to a very wide range of problems, is not straightforward to determine an appropriate model neural network architecture and the corresponding set of model weights required to achieve a useful approximator. In this paper we examine the challenge of finding the optimal weights for a given neural network model architecture by comparing the performance of several well known optimisation algorithms, Adam~\cite{Kingma2017}, LM ~\cite{Levenberg1944} ~\cite{Marquardt1963}, BFGS~\cite{Nocedal2006} and BFGS-L~\cite{Liu1989}. We apply these algorithms to a well known but challenging problem where the frequency is fixed and the amplitude varies. We also investigate the process of model optimisation by studying the fit achieved in incremental steps.

Finally, the results presented here have significant practical implications. This study was motivated by earlier studies where we examined applying machine learning (ML) to detecting faults of fielded machinery using unsupervised learning ~\cite{Wang2021a} ~\cite{Wang2021} where the function approximation and the choice of the optimizer was a key task in delivering breakthrough results. 

\section{Methods}

There are many quasi-Newton based optimisation algorithms available for training neural networks. All algorithms are a compromise between accuracy, robustness, computational efficiency, memory usage, scalability and flexibility. We have selected four very well known optimisation algorithms that have seen widespread application for our study, briefly summarized as follows.

\subsection{Adam}

The Adam algorithm~\cite{Kingma2017} is a first-order gradient based optimization of stochastic loss functions and has enjoyed widespread application to fitting large and complex deep neural networks. The method is robust, computationally efficient and works well on deep neural networks with millions of parameters. Tuning of the learning rate, learning rate warmup, learning rate schedules, momentum options, selection of regularizers and their parameters is needed to achieve the best optimisation result. We use the Adam optimiser algorithm as implemented in TensorFlow 2.

\subsection{BFGS}

The BFGS algorithm ~\cite{Nocedal2006} is a second-order iterative gradient method that computes the Hessian matrix of the loss function using gradient evaluations, ${\displaystyle {\mathcal {O}}(n^{2})}$, where $n$ is the number of parameters. Updates to the BFGS curvature matrix do not need a matrix inversion which reduces the computational cost significantly. As the Hessian matrix is the basis of the BFGS algorithm, memory usage grows as the square of the number of parameters which results in rapid growth in memory usage, making the approach unsuitable for neural networks with large numbers of parameters. The selection of BFGS algorithm parameters is straightforward. We use the BFGS optimiser as implemented in TensorFlow Probability~\cite{abadi2015}~\cite{TensorFlow2021}.

\subsection{L-BFGS}

L-BFGS~\cite{Liu1989} addresses the problem of large memory usage by the BFGS algorithm by saving only a few vectors that represent an estimate of the full Hessian matrix. L-BFGS is computationally more efficient than BFGS, uses less memory and can be applied to problems with larger numbers of parameters than BFGS. We use the L-BFGS optimiser as implemented in TensorFlow Probability~\cite{abadi2015}~\cite{TensorFlow2021}.

Both BFGS and L-BFGS can fall into local minima. By combining Adam as an initial optimisation algorithm, run over many hundreds of epochs, then continuing with BFGS and/or L-BFGS using the model weights delivered by Adam, the risk of falling into local minima can be reduced. This approach has been used in physics-Informed machine learning in order to solve systems of partial differential equations(PDEs)~\cite{Karniadakis2021}, for example, and is also used in this study.

\subsection{Levenberg-Marquardt}

Levenberg~\cite{Levenberg1944} was the original developer of this method for non-linear least squares based on the finding that simple gradient descent and Gauss-Newton iteration are complementary methods. Marquardt~\cite{Marquardt1963} extended the original method by scaling each component of the gradient according to the curvature by including the diagonal of the Hessian. Ananth~\cite{Ranganathan2004} provides an intuitive summary of the method. The LM method does require a matrix inversion which is computationally expensive and consumes significant amounts of memory, limiting the application of LM to moderately sized models with thousands of parameters. It is worth noting that modern computing hardware, with substantial computational capability and memory, has expanded the size of models to which LM can usefully be applied. As there is no built-in implementation of LM in Tensorflow and Keras, we imported a user contributed code for Levenberg-Marquardt (LM) algorithm by Fabio Di Marco~\cite{FabioDiMarco2021}.

\subsection{Test model}

Our test model is the function $y = sinc(10x)$ over the domain [-1.5, 1.5]. We build a neural network model consisting of an input layer, 2 Dense layers each with 20 hidden units, and an output layer with a total of 481 trainable parameters. We use the $tanh$ activation function in the two dense layers. In all cases we minimize the mean square error loss. We generate a training data set consisting of 20,000 evenly spaced points. We implemented the model in Python using Keras ~\cite{Chollet2015} and TensorFlow 2~\cite{abadi2015}.

\subsection{PINN model}

Physics Informed Neural Networks (PINNs) are neural networks designed to solve a variety of computational problems while accounting for the physical equations which govern their respective natural phenomena. PINN is an emerging deep learning technique that includes linear/nonlinear Ordinary Differential Equations (ODEs), Partial Differential Equations (PDEs) into the loss function of the NNs via automatic differentiation, in addition to any collected data, essentially acting as an approximate differential equation solver which may function well even with little to no training data. The PINN approach was originally developed by ~\cite{Raissi2017}, with two main applications in mind: data-driven solution (solving differential equations given data/initial conditions/equation parameters) and data-driven discovery of PDEs (find the parameters of the governing PDEs given collected data), by using the fact that deep neural networks can act as universal function approximators ~\cite{Hornik1989}. 

PINN’s initial design philosophy addressed the issue that for certain complex physical systems the collection of data may be restrictive. PINNs shine as they do not require huge data sets to produce a sufficiently good result. In most machine learning techniques, more training data means more accurate and robust results, and a sufficient amount of training data is needed for the models to produce anything of quality. The creators of PINNs took advantage of the abundance of domain specific information surrounding physical systems which may not be included in standard neural network methods of modelling. PINNs maximise the available knowledge of a problem by including already known mathematical/physical descriptions of the problems, in addition to the available training data, into the learning algorithms. Doing so may induce faster convergence, and result in a much more generalisable solution when compared to traditional machine learning methods, even with little to no training data ~\cite{Raissi2019a}.

This module implements a PINN model for Burgers' equation~\cite{Raissi2019a}. Burgers' equation is given by:
\begin{equation}
\frac{\partial u}{\partial t} + u \frac{\partial u}{\partial x} = \nu\frac{\partial^2 u}{\partial x^2}
\end{equation}
where $\nu$ is the kinematic viscosity, $x$ is the spatial coordinate, $t$ is the temporal coordinate, and $u(x,t)$ is the velocity of the fluid at point $(x,t)$. This is a simplification of Navier-Stokes equation where the velocity is in one spatial dimension and where external forces and pressure gradient terms are ignored. We apply the initial conditions $u(x, t=0) = -sin(x)$ and the boundary conditions $u(x=-1,+1, t) = 0.$ The PINN model predicts $u(x, t)$ at the point $(x, t)$.  $u(x, t)$ represents a typical engineering problem where abrupt changes are expected, which can be challenging for optimizers

We use an implementation of the solution for Burgers' equation~\cite{Raissi2019a} written in TensorFlow provided by ~\cite{okada39}. We compare solutions of the PINN model obtained using the L-BFGS-B and LM optimizers.

\section{Results and Discussion}

We first explored the performance of the Adam optimiser by allowing the model to train for a fixed number of epochs, at which point training was paused and the solution evaluated. The model training was then restarted using the model weights obtained when training was paused. In this way we can inspect the performance of the optimiser as training proceeds.

\begin{figure}[h!]
\begin{center}
\includegraphics[width=10.5cm, angle=-90]{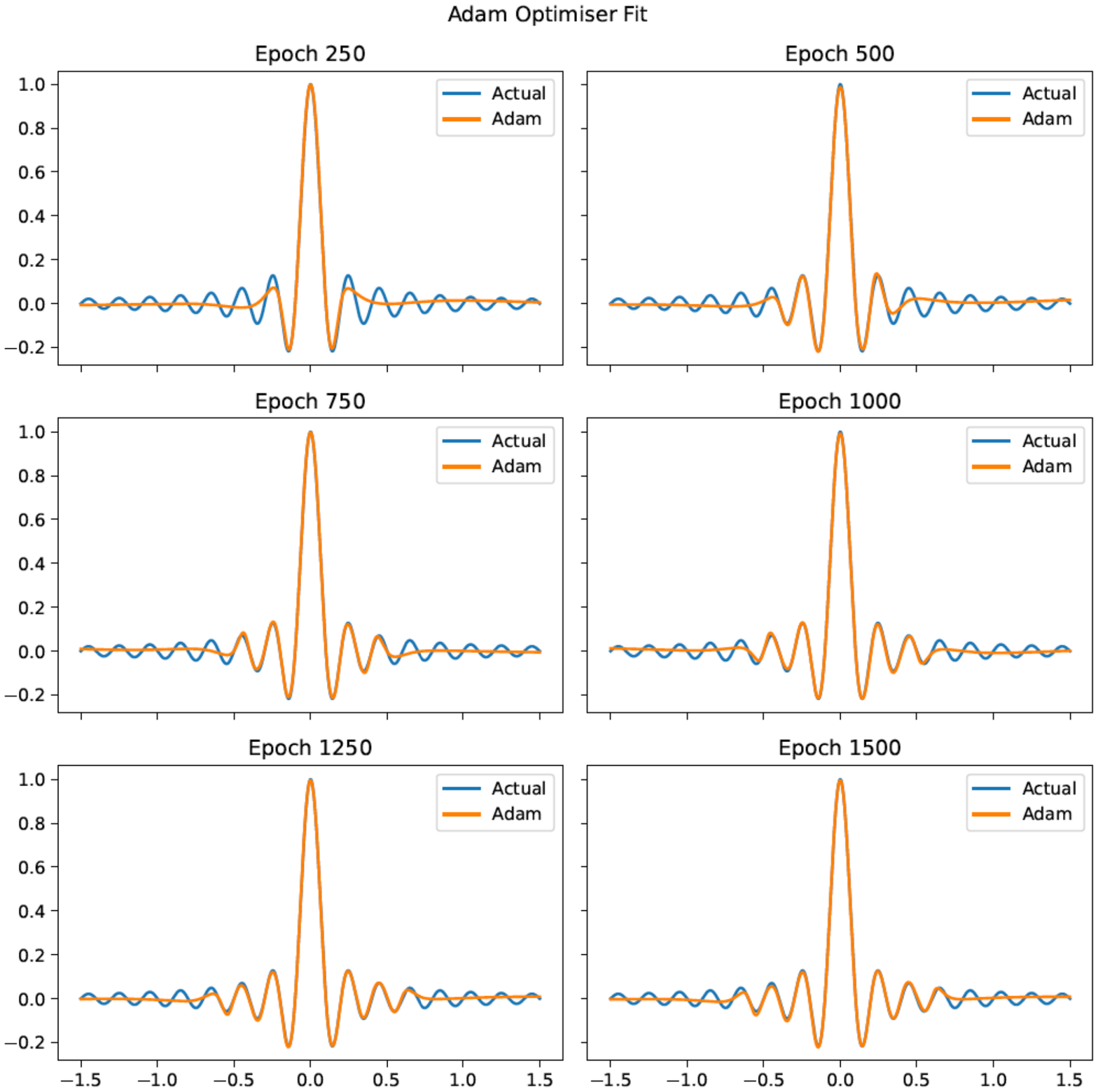}
\end{center}
\caption{The fit achieved to the function $y = sinc(10x)$ of the Adam optimizer after 250, 500, 750, 1000, 1250 and 1500 epochs. As expected the model fit improves as we increase the number of epochs. The Adam optimizer fits the higher amplitude oscillations first, then fits each subsequent lower amplitude oscillation in the function. Note that the frequency of the oscillations is constant and after 1500 epochs the model fit does not improve.}\label{fig:1}
\end{figure}

Figure ~\ref{fig:1} shows the fit achieved to the function $y = sinc(10x)$ of the Adam optimizer after 250, 500, 750,1000, 1250 and 1500 epochs. As expected the model fit improves as we increase the number of epochs. The Adam optimizer fits the higher amplitude oscillations first, then fits each subsequent lower amplitude oscillation in the function. Note that the frequency of the oscillations is constant i.e. it is the amplitude only that varies. After 1500 epochs we found that the model fit does not improve and from inspection of the graph we can see that the model has not achieved a highly accurate fit.  Simply relying on having obtained a minimum value for the MSE loss function cannot be relied on to ensure that we have obtained a satisfactory fit. In the following experiment we compare the fit obtained from multiple loss functions.

\begin{figure}[h!]
\begin{center}
\includegraphics[width=12.5cm]{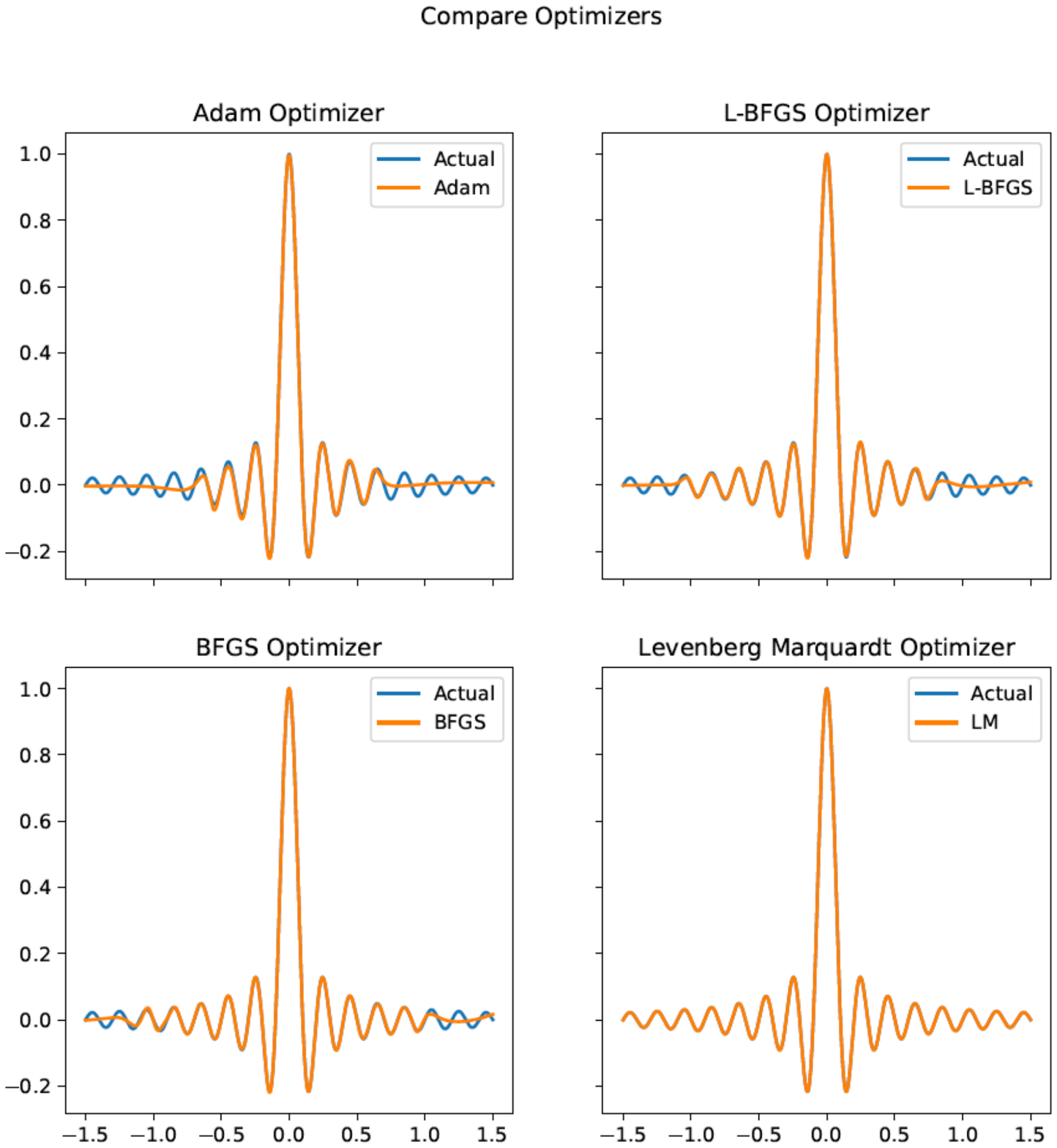}
\end{center}
\caption{A comparison of the performance of the Adam optimizer, an algorithm for first-order gradient-based optimisation that uses adaptive momentum, the Levenberg-Marquardt (LM) algorithm a second order method, BFGS a second order method and BFGS-L, a low memory version of BFGS where we fit the function $y = sinc(10x)$ using the same neural network. All model runs show the results of the best fit achieved. The Adam, L-BFGS and BFGS Optimizers have difficulty fitting the lower amplitude tails of the function.}\label{fig:2}
\end{figure}

Figure ~\ref{fig:2} provides a comparison of the performance of the Adam optimizer, the LM, BFGS and BFGS-L optimizers, where we fit the function $y = sinc(10x)$ using the identical neural network, settings and input data for all training runs. BFGS and L-BFGS use the weights generated by the Adam optimizer as their starting point as suggested in previous studies ~\cite{Raissi2019a}. All model runs show the results of the best fit achieved for each optimiser. The Adam, L-BFGS and BFGS optimizers have difficulty fitting the lower amplitude tails of the function with the LM model achieving the best fit. For Adam, the minimum value of MSE loss function was 0.0002018958 and occurred at optimisation step 1,464, for BFGS-L, 8.655213e-05 at step 14,558, for BFGS, 4.2915726e-06 at step 12,477 and for LM, 1.642267e-07 at step 150, the maximum number of epochs set for LM model training. The LM MSE value is $\approx$26x lower than that obtained by BFGS, $\approx$527x lower than L-BFGS and $\approx$1,230x lower than Adam. The compute time for each optimisation method, measured using the Python time module, was for Adam 76s, BFGS 58s, L-BFGS 172 s and LM 116s, noting that the total time for BFGS and L-BFGS optimisation includes Adam as the starting point, results in a total time for BFGS of 134s and for L-BFGS of 248s.

\begin{figure}[h!]
\begin{center}
\includegraphics[width=11.5cm, angle=-90]{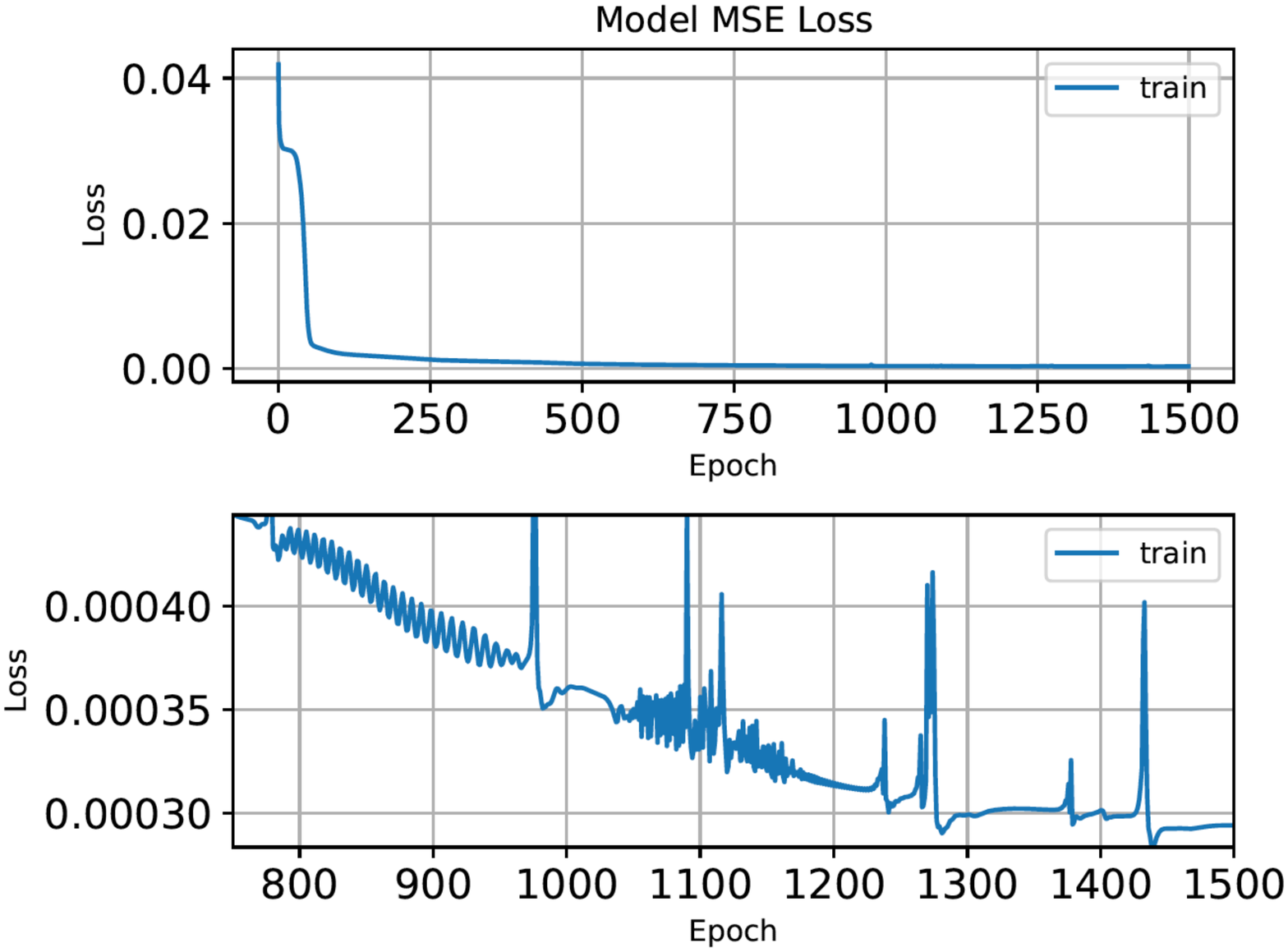}
\end{center}
\caption{ The MSE loss value, plotted on a linear scale, using the Adam optimizer for each training epoch where we fit the function $y = sinc(10x)$ using a neural network. The upper panel plots the MSE loss over the entire training run with the bottom panel focussing on the end of the model run. Note that the scale of the two plots is not the same. }\label{fig:3}
\end{figure}

Figure ~\ref{fig:3}  shows MSE loss value, plotted on a linear scale, using the Adam optimizer for each training epoch where we fit the function $y = sinc(10x)$ using a neural network. The upper panel plots the MSE loss over the entire training run with the bottom panel focussing on the end of the model run. The Adam optimiser exhibits significant variability between epochs before reaching a minimum value after $\approx$1300 epochs.

\begin{figure}[h!]
\begin{center}
\includegraphics[width=11.5cm, angle=-90]{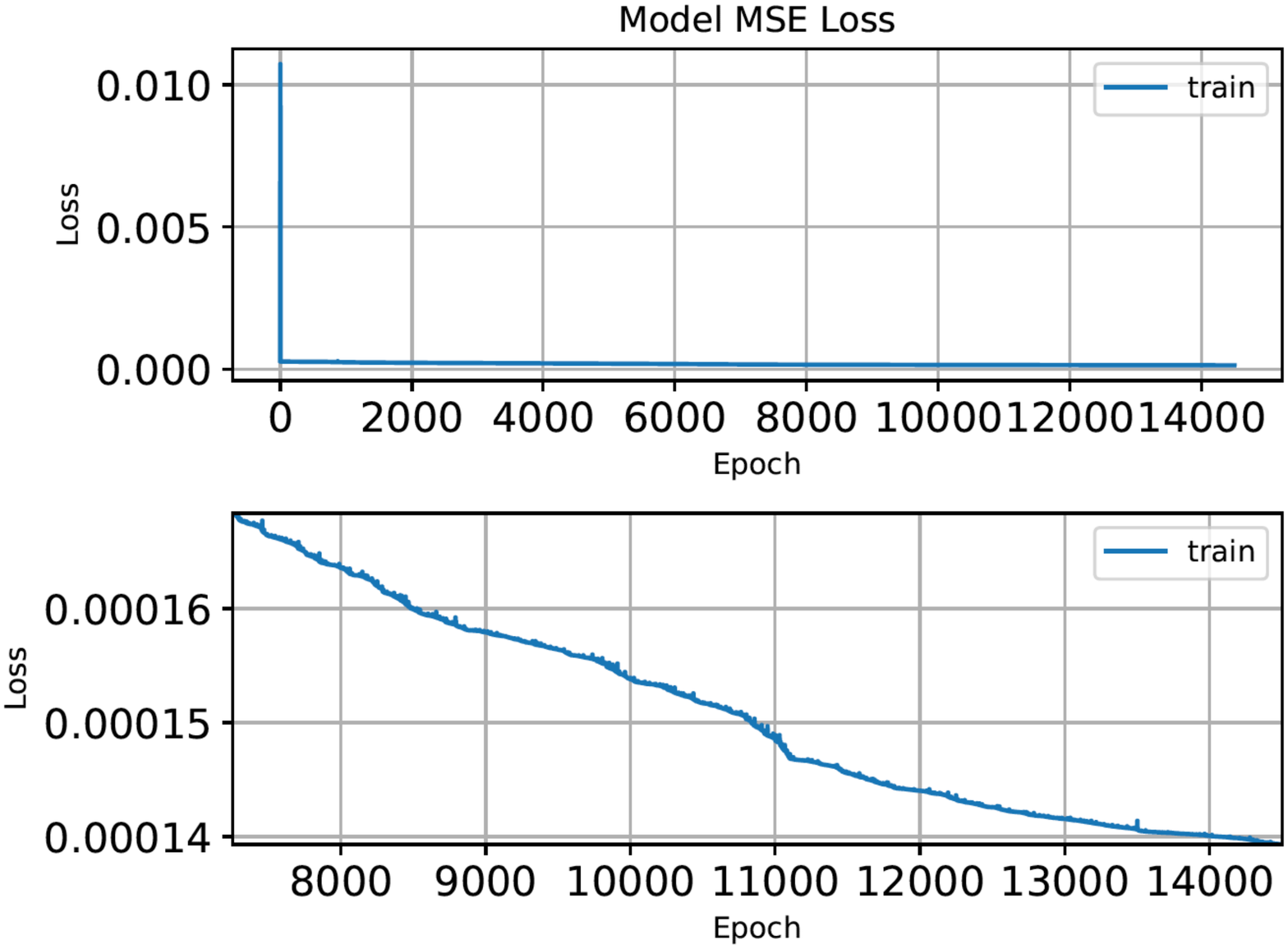}
\end{center}
\caption{The MSE loss value, plotted on a linear scale, using the L-BFGS optimizer for each training epoch where we fit the function $y = sinc(10x)$ using a neural network. The upper panel plots the MSE loss over the entire training run with the bottom panel focussing on the end of the model run. Note that the scale of the two plots is not the same.}\label{fig:4}
\end{figure}

\begin{figure}[h!]
\begin{center}
\includegraphics[width=11.5cm, angle=-90]{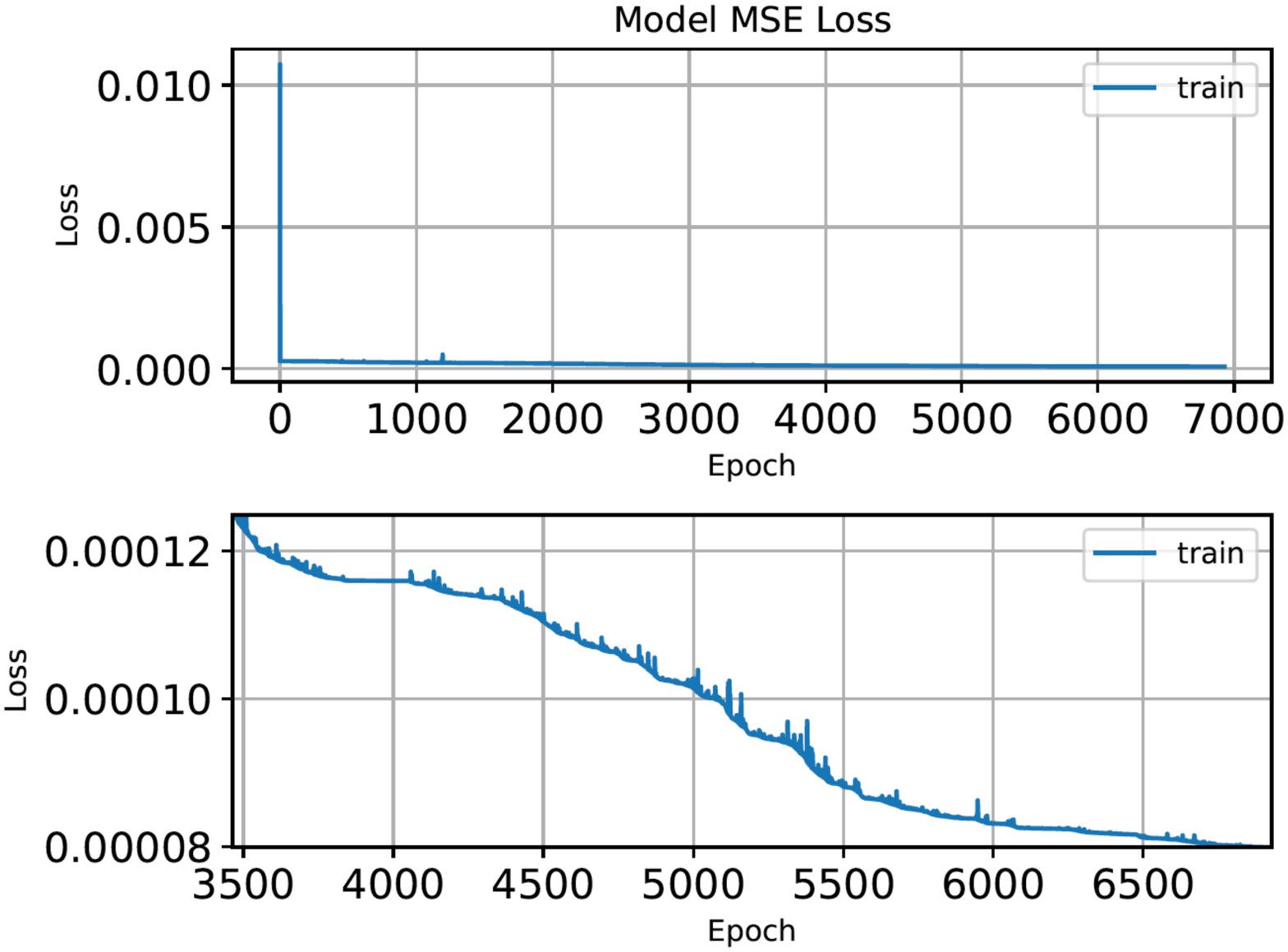}
\end{center}
\caption{ The MSE loss value, plotted on a linear scale, using the BFGS optimizer for each training epoch where we fit the function $y = sinc(10x)$ using a neural network. The upper panel plots the MSE loss over the entire training run with the bottom panel focussing on the end of the model run. Note that the scale of the two plots is not the same. }\label{fig:5}
\end{figure}

Figure ~\ref{fig:4} and Figure ~\ref{fig:5} compare the MSE loss value, plotted on a linear scale, using the L-BFGS and BFGS optimizers. Note that the L-BFGS and BFGS optimizers were initialised with the set of weights from the conclusion of the Adam optimzer run. Figure 4 shows that the L-BFGS optimiser takes many more steps before converging at a higher value of the MSE than BFGS. This would be expected given that L-BFGS does not calculate the full Hessian matrix. Where parameter numbers are in the low to medium range BFGS is likely the better choice of optimizer than L-BFGS. Only where memory usage or computational time becomes too large would L-BFGS be recommended.

\begin{figure}[h!]
\begin{center}
\includegraphics[width=11.5cm, angle=-90]{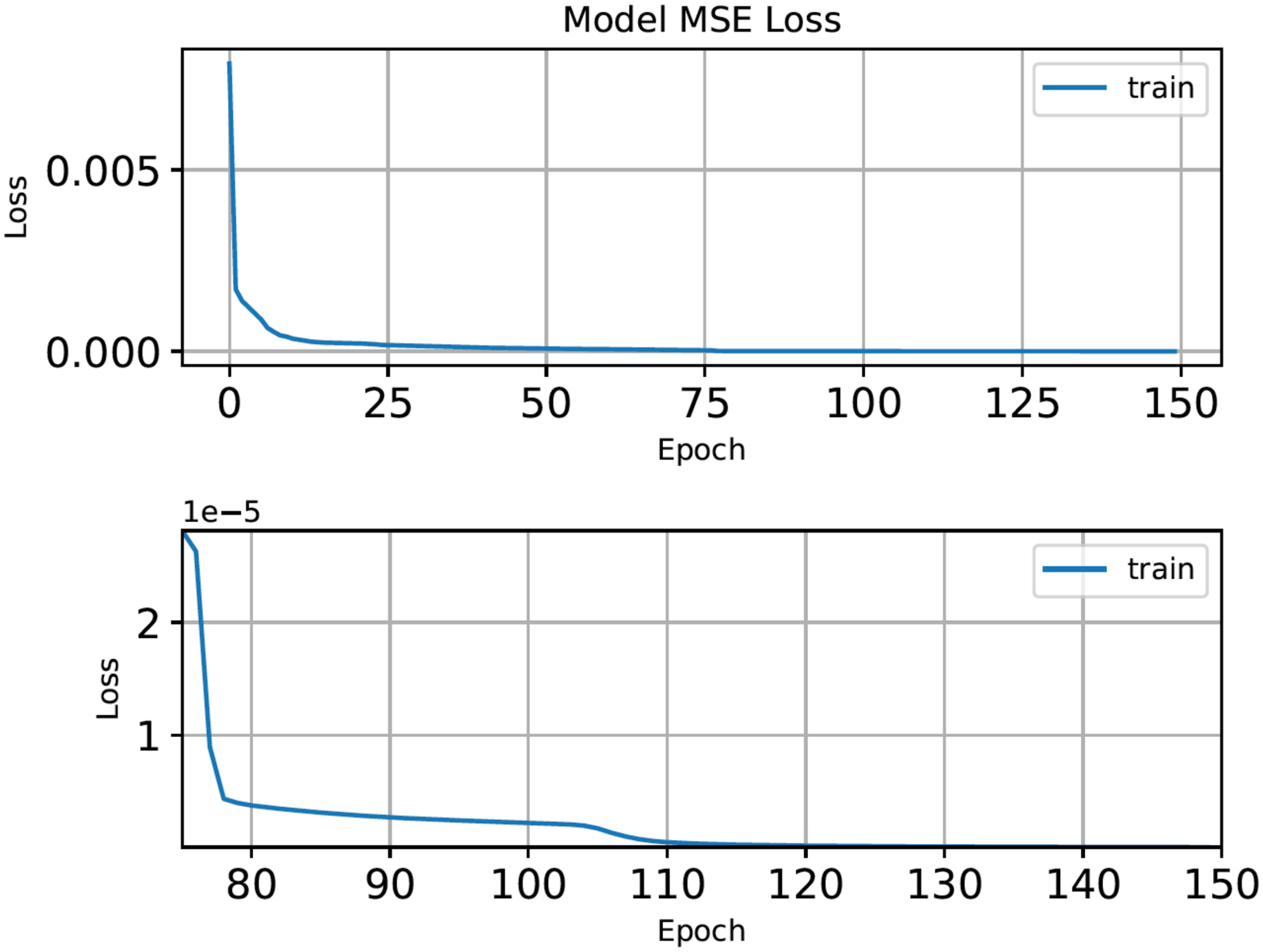}
\end{center}
\caption{  The MSE loss value, plotted on a linear scale, using the LM optimizer for each training epoch where we fit the function $y = sinc(10x)$ using a neural network. The upper panel plots the MSE loss over the entire training run with the bottom panel focussing on the end of the model run. Note that the scale of the two plots is not the same.}\label{fig:6}
\end{figure}

Figure ~\ref{fig:6} presents the MSE loss value, plotted on a linear scale, using the LM optimiser. The LM optimizer reduces the MSE smoothly and in only a few epochs. The MSE loss value rapidly falls below the Adam, L-BFGS and BFGS optimizer MSE values. The benefits of the high quality model fit achieved by the LM optimizer are clearly demonstrated in Figure ~\ref{fig:2} where, in contrast with the other optimizers, the low amplitude (not frequency) components of the function are well fitted.

\begin{figure}[h!]
\begin{center}
\includegraphics[width=11.5cm, angle=-90]{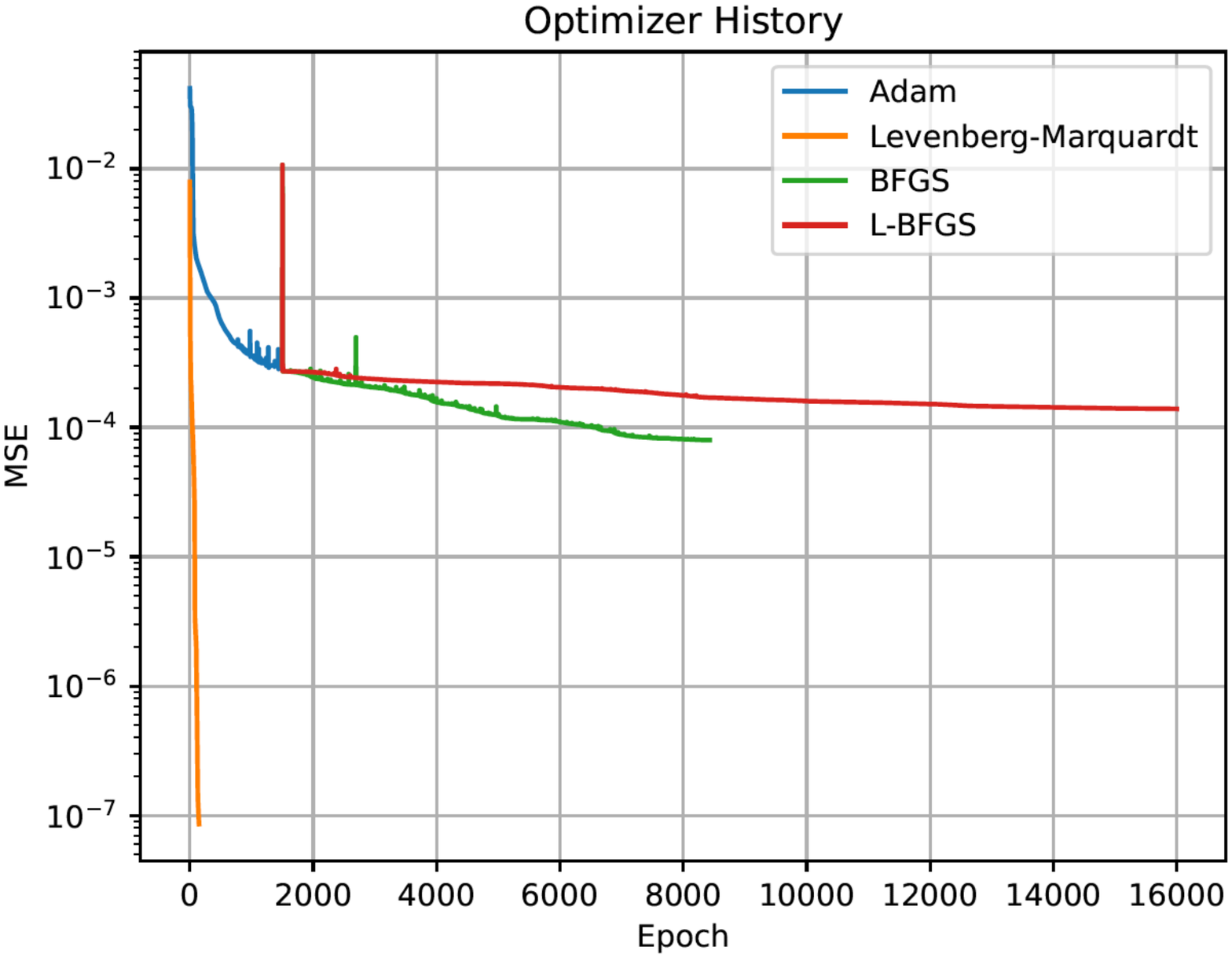}
\end{center}
\caption{A comparison of the performance of the Adam optimizer, an algorithm for first-order gradient-based optimisation that uses adaptive momentum, the Levenberg-Marquardt (LM) algorithm a second order method, BFGS a second order method and BFGS-L, a low memory version of BFGS, all plotted on a log scale. The BFGS and L-BFGS algorithms commence fitting using the final state of the Adam optimizer after the Adam optimizer has completed 1500 time steps. The Levenberg-Marquardt (LM) algorithm achieves an improved MSE loss value compared to the Adam optimizer after 10 epochs and the BFGS algorithm after 20 epochs.}\label{fig:7}
\end{figure}

Figure ~\ref{fig:7} compares the results obtained using all optimizers plotted on a log scale. The log scale makes it possible to more easily compare optimiser performance. Figure ~\ref{fig:7} clearly demonstrates the superior performance of the LM method with a rapid minimization of the MSE to levels several orders of magnitude lower than that achieved by other methods. The LM optimizer requires no tuning to produce these results. While these results were obtained using a simple test function, earlier studies where we examined applying machine learning (ML) to detecting faults of fielded machinery using unsupervised learning ~\cite{Wang2021a} ~\cite{Wang2021} ~\cite{Wang2021b}with noisy data collected from aircraft sensors, support the conclusion that the LM optimizer can deliver superior performance in a real-world setting.

Having determined that the LM optimizer can deliver improved performance on problems with a small to medium number number of weights we have also considered the application of the LM optimizer to PINNs, an important new class of NN applications. Many PINN models require a small to medium number of weights and would benefit from the high accuracy solutions, often required of science and engineering problems, that the LM model is able to deliver. 

\begin{figure}[h!]
\begin{center}
\includegraphics[width=15.5cm]{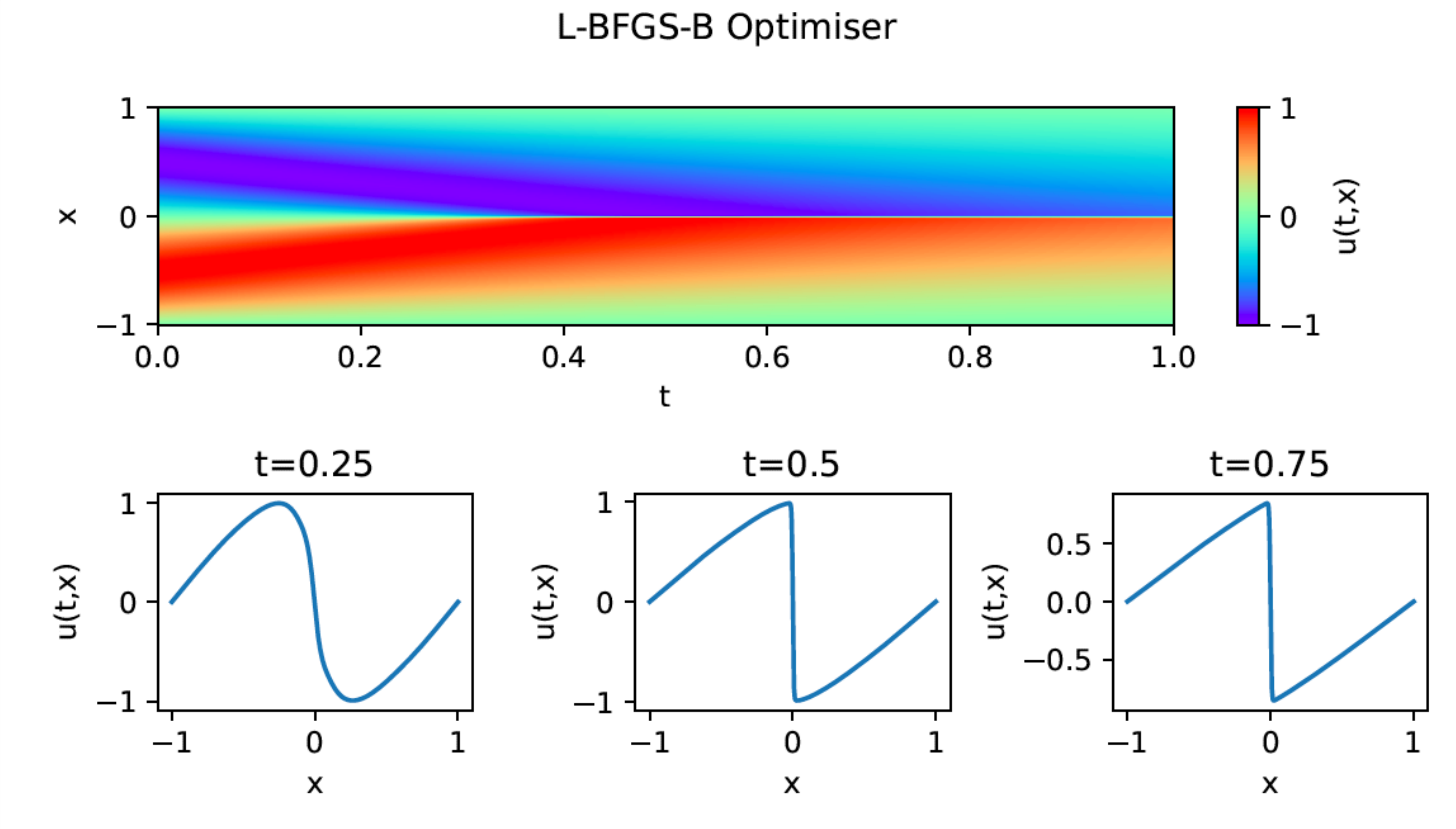}
\end{center}
\caption{Solution of the Burgers equation solved using a PINN model using the L-BFGS-B optimizer.}\label{fig:8}
\end{figure}

\begin{figure}[h!]
\begin{center}
\includegraphics[width=15.5cm]{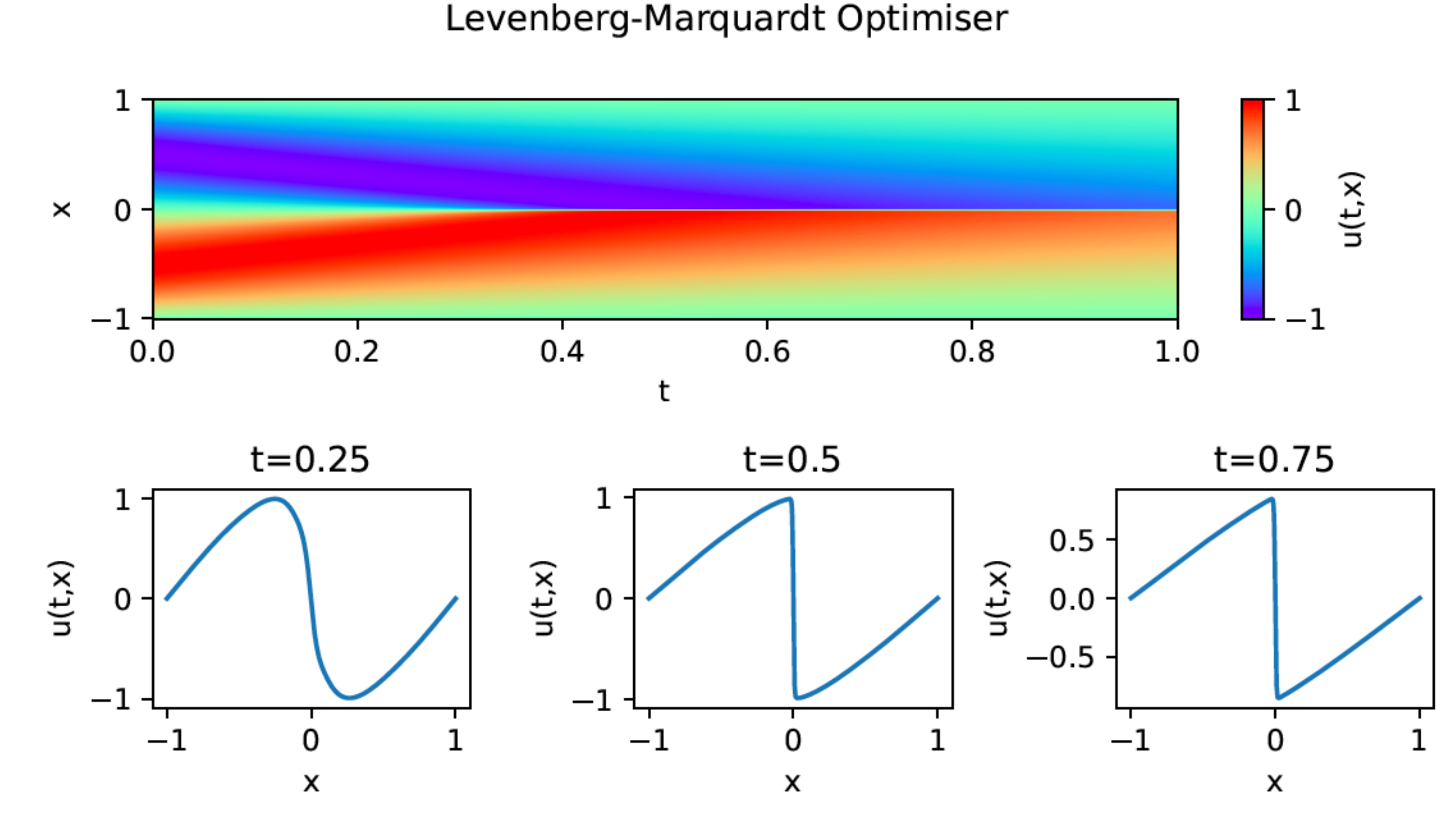}
\end{center}
\caption{Solution of the Burgers equation solved using a PINN model using the Lavenburg-Marquardt optimizer. The solution matches that obtained using the L-BFGS-B optimizer.}\label{fig:9}
\end{figure}

\begin{figure}[h!]
\begin{center}
\includegraphics[width=11.5cm, angle=-90]{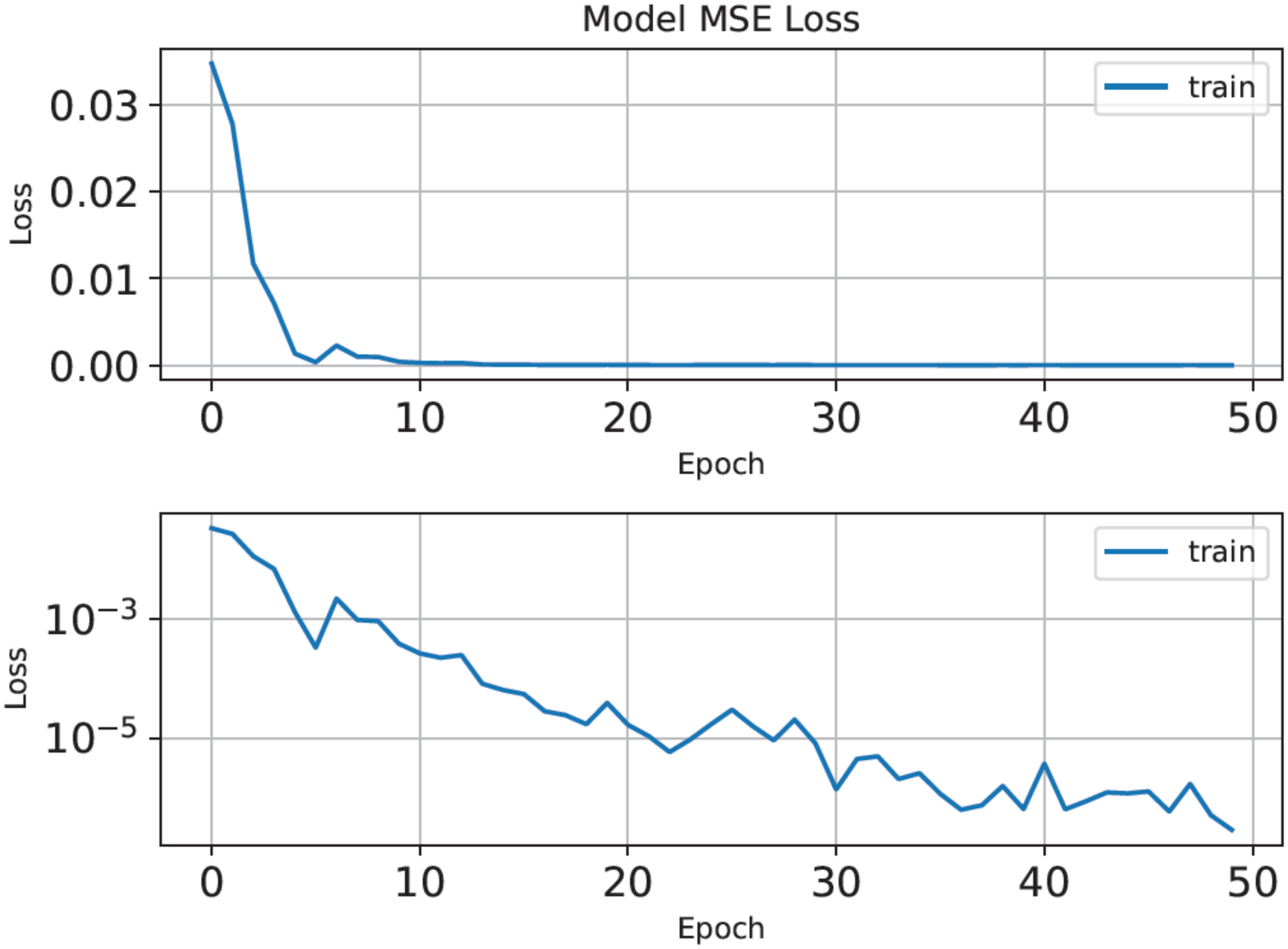}
\end{center}
\caption{The MSE loss as a function of epoch using the LM optimizer for the Burgers equation solved using a PINN model. The upper panel shows the loss on a linear scale with the lower panel showing the loss on a log scale.  The final loss value is lower than that obtained in previous studies.}\label{fig:10}
\end{figure}

Figure ~\ref{fig:8} presents the results of fitting a PINN model, as described above, to Burgers' equation~\cite{Raissi2019a} using the L-BFGS-B optimizer. This version of the BFGS model extends L-BFGS to take into account the box constraints needed to solve Burgers' equation. The results obtained are identical to that obtained previously ~\cite{Raissi2019a}. Figure ~\ref{fig:9} presents the results obtained using the LM optimizser and are identical to the results obtained using L-BFGS-B. Finally, Figure ~\ref{fig:10} shows the MSE loss curve for the LM optimiser over 50 epochs, with the MSE value approaching \num{1e-7}, several orders of magnitude lower than previously reported MSE values, with the potential for further improvement. The application of the LM model required no hyperparameter tuning.

The previous figures all show results with the model architecture fixed for all optimizers. As the Adam optimizer is widely used when fitting deep learning models, we have also investigated how to obtain the best fit to the test function using only the Adam optimizer by modifying the model architecture.  As previously shown, the Adam optimizer delivered an MSE value approximately 3 orders of magnitude higher than the LM optimizer. To achieve an improved result for the Adam optimizer we investigated a range of hidden units (HU) using 16,32,48,64 and 80 HU and varied the number of dense layers in the model from 1-4. We use the same number of hidden units for each layer. Using this basic grid search, we have a total of 20 model fit runs using a combination of 5 hidden units and 4 layer combinations. We fit all models over 5000 epochs.

\begin{figure}[h!]
\begin{center}
\includegraphics[width=11.5cm, angle=-90]{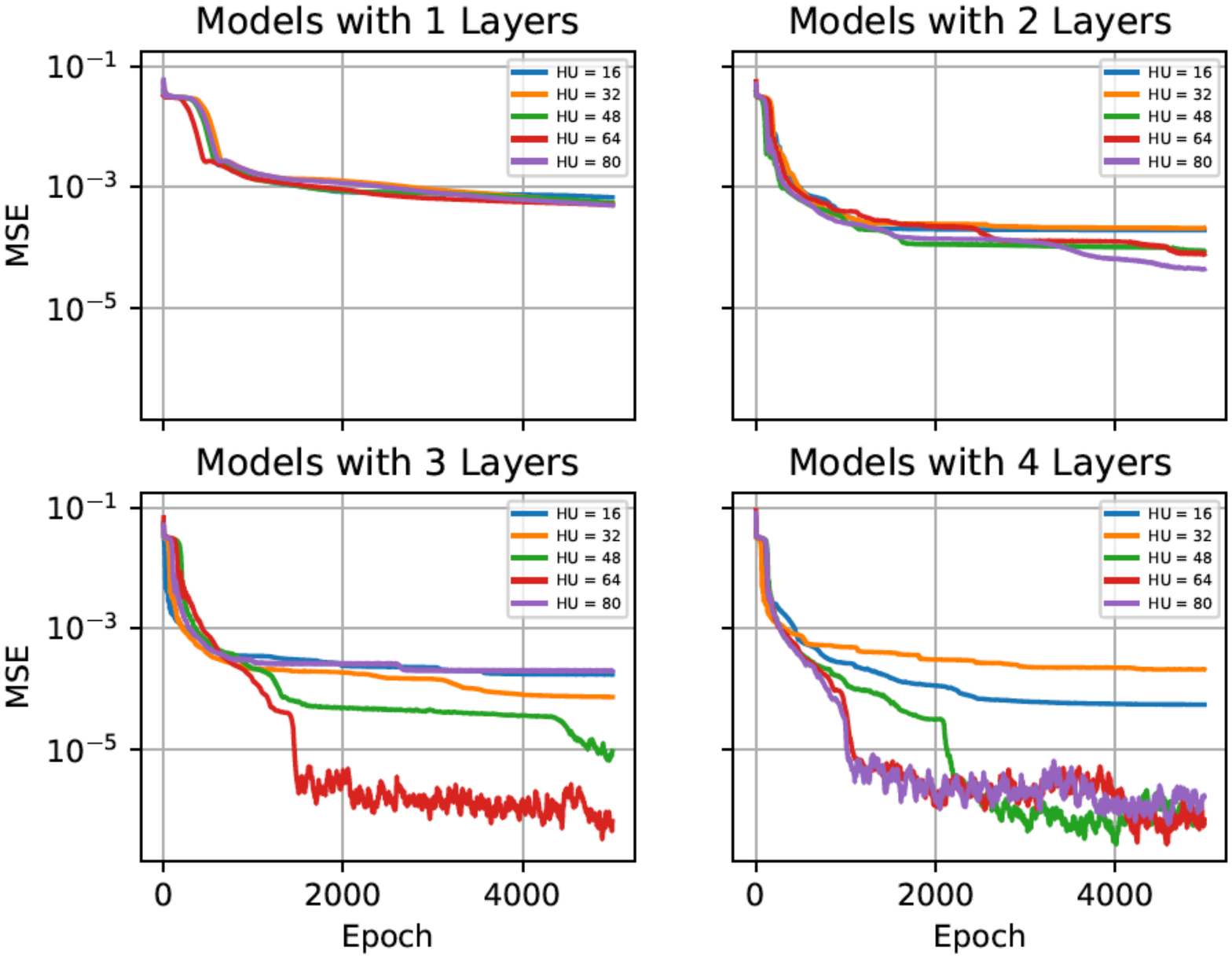}
\end{center}
\caption{The MSE loss as a function of epoch using the Adam optimizer for the test function over 5000 epochs. Each panel shows the results with a model with 1, 2, 3 or 4 layers with the number of hidden units in all layers of the model equal to 16, 32, 48, 64 or 80. Note that some curves are hidden beneath other results.}.\label{fig:11}
\end{figure}

\begin{figure}[h!]
\begin{center}
\includegraphics[width=11.5cm, angle=-90]{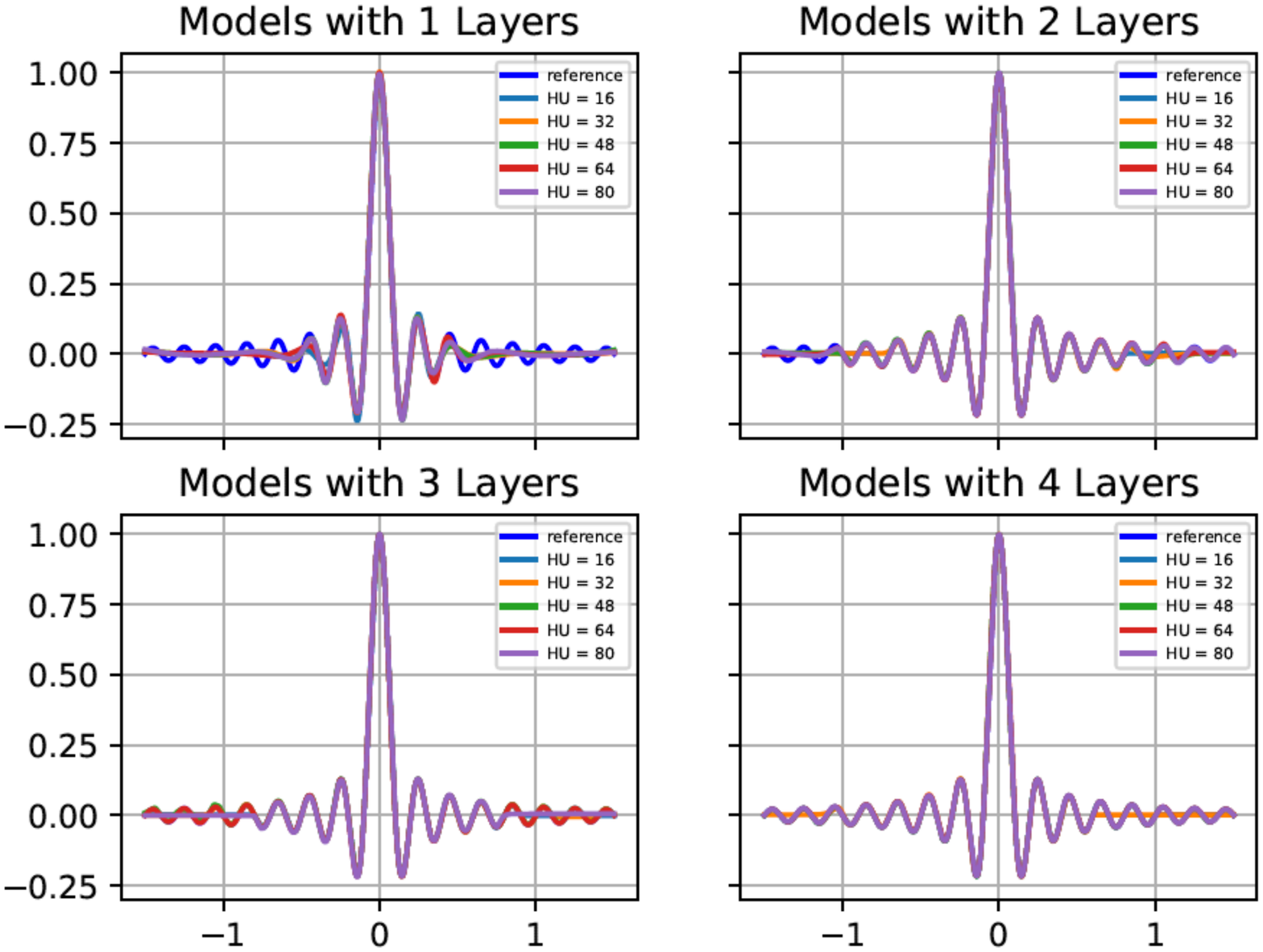}
\end{center}
\caption{The model fit achieved using the Adam optimiser using a model with 1, 2, 3 or 4 layers with the number of hidden units in all layers of the model equal to 16, 32, 48, 64 or 80. Note that some curves are hidden beneath other results.}\label{fig:12}
\end{figure}

Figure ~\ref{fig:11} shows the results of this grid search organised as four panels with an increasing number of model layers with each panel including the model fit for all five HU values. Using only a single layer the model fit is very similar across all HU values and does not improve on the MSE of the original model. The maximum number of parameters using a 1 layer model with 80 HUs is 241 compared with the 481 parameters in the original test function with 2 layers and 20 HUs. With the 2 layer model all models achieve improved performance compared with the single layer models. The model with 80 HUs and 2 layers (6,721 parameters) achieves the best result with the MSE value falling below \num{1e-4}, an improvement on the original model. The model with 3 layers further improves on these results, however the model with 64 HUs (8,513 parameters), rather than 80 HUs (13,201 parameters), produces the best results with the MSE value dropping below \num{1e-6}. With 4 layers the model is able to achieve good model results 48, 64, and 80 HUs with the best results at 48 HUs (7,201 parameters) and 64 HUs (12,673 parameters). The models with the larger number of parameters were approaching the total number of training samples (20,000). Figure ~\ref{fig:12} shows the corresponding model fit achieved using the models in Figure ~\ref{fig:11}. The model fit is consistent with the MSE values achieved during model fitting.

Figure ~\ref{fig:11} and ~\ref{fig:12} illustrate that the Adam optimizer performs best on this problem primarily by increasing the number of layers and then by increasing the number of HUs. Note that increasing the number of HUs to 80 did not produce the best model for models with 3 and 4 layers. The best Adam optimizer results required $\approx$15x more parameters with a 3 layer model and $\approx$26x more parameters with 4 layer model as the LM optimized model. While the Adam optimiser results can be improved, even with this much greater number of model parameters, the Adam model cannot outperform the LM optimiser. This result implies that Adam optimiser may require deeper models with large numbers of hidden units in order to achieve the best model fit. The LM optimiser results illustrate that it may be possible build models with far fewer parameters.

\section{Conclusions}

We compared a range of well known optimizers applied to fitting neural networks with a small to medium number of weights. We studied the performance of Adam, the Levenberg-Marquardt (LM) algorithm, BFGS and L-BFGS and concluded that the LM optimiser delivered significant advantages over the other methods, including orders of magnitude improvement in the MSE values, rapid optimisation and straightforward application as no hyperparameter tuning was required. 

Using these Optimizers we fit the function $y = sinc(10x)$ using a neural network with a few parameters. This function has a variable amplitude and a constant frequency. We observed that as the model fit progressed, the higher amplitude components of the function were fit first. We found that only the LM optimiser could fit this function well across the full range of amplitudes.

We have also demonstrated the usefulness of the LM optimiser to PINNs by solving the Burgers equation to a higher accuracy than has previously been achieved. Combined with the observation that the LM method can achieve an accurate fit where other optimizers cannot, the LM method is likely to be very useful for PINNs and to the broader class of NN models with a small to medium number of parameters.  

It would likely benefit the broader ML community if the LM and BFGS optimizers included in could be made available without additional wrapping in TensorFlow.  This is recommended as there are a large number of potential applications that would benefit significantly from the ready availability of these optimizers.

\bibliographystyle{chicago}
\bibliography{Optimiser2022.bib}

\begin{thebibliography}{}

\bibitem[\protect\citeauthoryear{Abadi, Agarwal, Barham, Brevdo, Chen, Citro,
  Corrado, Davis, Dean, Devin, Ghemawat, Goodfellow, Harp, Irving, Isard,
  Jozefowicz, Jia, Kaiser, Kudlur, Levenberg, Man{\'{e}}, Schuster, Monga,
  Moore, Murray, Olah, Shlens, Steiner, Sutskever, Talwar, Tucker, Vanhoucke,
  Vasudevan, Vi{\'{e}}gas, Vinyals, Warden, Wattenberg, Wicke, Yu, and
  Zheng}{Abadi et~al.}{2015}]{abadi2015}
Abadi, M., A.~Agarwal, P.~Barham, E.~Brevdo, Z.~Chen, C.~Citro, G.~S. Corrado,
  A.~Davis, J.~Dean, M.~Devin, S.~Ghemawat, I.~Goodfellow, A.~Harp, G.~Irving,
  M.~Isard, R.~Jozefowicz, Y.~Jia, L.~Kaiser, M.~Kudlur, J.~Levenberg,
  D.~Man{\'{e}}, M.~Schuster, R.~Monga, S.~Moore, D.~Murray, C.~Olah,
  J.~Shlens, B.~Steiner, I.~Sutskever, K.~Talwar, P.~Tucker, V.~Vanhoucke,
  V.~Vasudevan, F.~Vi{\'{e}}gas, O.~Vinyals, P.~Warden, M.~Wattenberg,
  M.~Wicke, Y.~Yu, and X.~Zheng (2015).
\newblock {TensorFlow: Large-scale machine learning on heterogeneous systems}.

\bibitem[\protect\citeauthoryear{Chollet}{Chollet}{2015}]{Chollet2015}
Chollet, F. (2015).
\newblock {Keras}.

\bibitem[\protect\citeauthoryear{{Fabio Di Marco}}{{Fabio Di
  Marco}}{2021}]{FabioDiMarco2021}
{Fabio Di Marco} (2021).
\newblock {Setup Levenberg-Marquardt optimization algorithm on TensorFlow}.

\bibitem[\protect\citeauthoryear{Hornik, Stinchcombe, and White}{Hornik
  et~al.}{1989}]{Hornik1989}
Hornik, K., M.~Stinchcombe, and H.~White (1989).
\newblock {Multilayer feedforward networks are universal approximators}.
\newblock {\em Neural Networks\/}~{\em 2\/}(5), 359--366.

\bibitem[\protect\citeauthoryear{Karniadakis, Kevrekidis, Lu, Perdikaris, Wang,
  and Yang}{Karniadakis et~al.}{2021}]{Karniadakis2021}
Karniadakis, G.~E., I.~G. Kevrekidis, L.~Lu, P.~Perdikaris, S.~Wang, and
  L.~Yang (2021).
\newblock {Physics-informed machine learning}.
\newblock {\em Nature Reviews Physics\/}~{\em 3\/}(6), 422--440.

\bibitem[\protect\citeauthoryear{Kingma and Ba}{Kingma and
  Ba}{2015}]{Kingma2017}
Kingma, D.~P. and J.~L. Ba (2015).
\newblock {Adam: A method for stochastic optimization}.
\newblock In {\em 3rd International Conference on Learning Representations,
  ICLR 2015 - Conference Track Proceedings}. ArxIV.

\bibitem[\protect\citeauthoryear{Lecun, Bengio, and Hinton}{Lecun
  et~al.}{2015}]{Lecun2015}
Lecun, Y., Y.~Bengio, and G.~Hinton (2015).
\newblock {Deep learning}.

\bibitem[\protect\citeauthoryear{Levenberg}{Levenberg}{1944}]{Levenberg1944}
Levenberg, K. (1944).
\newblock {A method for the solution of certain non-linear problems in least
  squares}.
\newblock {\em Quarterly of Applied Mathematics\/}~{\em 2\/}(2), 164--168.

\bibitem[\protect\citeauthoryear{Liu and Nocedal}{Liu and
  Nocedal}{1989}]{Liu1989}
Liu, D.~C. and J.~Nocedal (1989).
\newblock {On the limited memory BFGS method for large scale optimization}.
\newblock {\em Mathematical Programming\/}~{\em 45\/}(1-3), 503--528.

\bibitem[\protect\citeauthoryear{Marquardt}{Marquardt}{1963}]{Marquardt1963}
Marquardt, D.~W. (1963).
\newblock {An Algorithm for Least-Squares Estimation of Nonlinear Parameters}.
\newblock {\em Journal of the Society for Industrial and Applied
  Mathematics\/}~{\em 11\/}(2), 431--441.

\bibitem[\protect\citeauthoryear{Nocedal and Wright}{Nocedal and
  Wright}{2006}]{Nocedal2006}
Nocedal, J. and S.~J. Wright (2006).
\newblock {Numerical optimization}.
\newblock In {\em Springer Series in Operations Research and Financial
  Engineering}, pp.\  1--664.

\bibitem[\protect\citeauthoryear{Okada39}{Okada39}{2022}]{okada39}
Okada39 (2022).
\newblock {Physics Informed Neural Network (PINN) for Burgers' equation.}

\bibitem[\protect\citeauthoryear{Raissi, Perdikaris, and Karniadakis}{Raissi
  et~al.}{2019}]{Raissi2019a}
Raissi, M., P.~Perdikaris, and G.~Karniadakis (2019, feb).
\newblock {Physics-informed neural networks: A deep learning framework for
  solving forward and inverse problems involving nonlinear partial differential
  equations}.
\newblock {\em Journal of Computational Physics\/}~{\em 378}, 686--707.

\bibitem[\protect\citeauthoryear{Raissi, Perdikaris, and Karniadakis}{Raissi
  et~al.}{2017}]{Raissi2017}
Raissi, M., P.~Perdikaris, and G.~E. Karniadakis (2017).
\newblock {Physics Informed Deep Learning (Part II): Data-driven Discovery of
  Nonlinear Partial Differential Equations}.
\newblock {\em arXiv preprint arXiv:1711.10561\/}.

\bibitem[\protect\citeauthoryear{Ranganathan}{Ranganathan}{2004}]{Ranganathan2004}
Ranganathan, A. (2004).
\newblock {The Levenberg-Marquardt Algorithm}.
\newblock {\em Internet httpexcelsior cs ucsb educoursescs290ipdfL MA
  pdf\/}~{\em 142\/}(June).

\bibitem[\protect\citeauthoryear{TensorFlow}{TensorFlow}{2021}]{TensorFlow2021}
TensorFlow (2021).
\newblock {TensorFlow Probability}.

\bibitem[\protect\citeauthoryear{Wang, Taylor, and Bala}{Wang
  et~al.}{2021}]{Wang2021b}
Wang, W., J.~Taylor, and B.~Bala (2021).
\newblock {Exploiting the Power of Levenberg-Marquardt Optimizer with Anomaly
  Detection in Time Series}.

\bibitem[\protect\citeauthoryear{Wang, Taylor, and Rees}{Wang
  et~al.}{2021a}]{Wang2021a}
Wang, W., J.~Taylor, and R.~J. Rees (2021a).
\newblock {Recent Advancement of Deep Learning Applications to Machine
  Condition Monitoring Part 1: A Critical Review}.

\bibitem[\protect\citeauthoryear{Wang, Taylor, and Rees}{Wang
  et~al.}{2021b}]{Wang2021}
Wang, W., J.~Taylor, and R.~J. Rees (2021b).
\newblock {Recent Advancement of Deep Learning Applications to Machine
  Condition Monitoring Part 2: Supplement Views and a Case Study}.
\newblock {\em Acoustics Australia\/}~{\em 49\/}(2), 221--228.

\end{thebibliography}

\pagebreak

\section*{Figures}

\

\end{document}